# Conscientious Classification:
## A Data Scientist's Guide to Discrimination-Aware Classification

**Authors:**

Brian d'Alessandro[1,2]

Cathy O'Neil[3]

Tom LaGatta[4]

[5]

## Abstract

Recent research has helped to cultivate growing awareness that machine learning systems fueled by big data can create or exacerbate troubling disparities in society. Much of this research comes from outside of the practicing data science community, leaving its members with little concrete guidance to proactively address these concerns. This article introduces issues of discrimination to the data science community on its own terms. In it, we tour the familiar data mining process while providing a taxonomy of common practices that have the potential to produce unintended discrimination. We also survey how discrimination is commonly measured, and suggest how familiar development processes can be augmented to mitigate systems' discriminatory potential. We advocate that data scientists should be intentional about modeling and reducing discriminatory outcomes. Without doing so, their efforts will result in perpetuating any systemic discrimination that may exist, but under a misleading veil of data-driven objectivity.

## 1. Introduction

For much of our industrial age, *automation* was a pursuit primarily focused on improving the productivity of low-skilled and repetitive manual tasks. In our digitized present, *automation* is increasingly moving from the hand to the brain. Even in its early years, machine learning (ML) systems successfully enabled firms to overcome human mental limitations, such as the consistency and speed of decision making, as well as understanding risk. Such innovations were a boon to financial services, where managing credit risk is the fundamental skill for a firm's survival, as well as to customers, many of whom found themselves newly able to fully participate in the economy.

[1] Department of Data Science, Zocdoc, New York, New York.

[2] NYU Center for Data Science, New York, New York.

[3] ORCAA, New York, New York.

[4] Splunk, New York, New York.

[5] Address correspondence to: Brian d'Alessandro, Department of Data Science, Zocdoc, New York, NY, E-mail: briand@gmail.com

Such benefits were happening when ML systems were meager: operating on single machine servers, scoring offline, and using simple linear models or decision trees with just tens of variables. Now that our data has grown by several orders of magnitude and learning can be distributed, online and with more depth, the range of applications and benefits have grown by orders of magnitude. That's the good news. The bad news is that as new applications are being tested, some subtle and ugly truths have also come to the fore.

While ML may be presented with an air of objectivity, it can also reinforce historical systemic biases. Data-driven decisions have the potential to negatively impact already disadvantaged populations, despite the fact that they might seem more fair. Such decisions can reinforce existing disparities, prompting concerns with 'disparate impact' in the broader debate about the impact of big data, ML, and artificial intelligence (AI)[1,2].

Disparate impact is not a new problem. Government regulators have long enforced legal restrictions to prevent discrimination, unintended or otherwise, in industries such as financial services[3]. But the proliferation of ML systems into new domains is complicating their job immensely. ML's "black box algorithms," which use datasets of enormous breadth and complexity, are themselves typically uninterpretable. Regulators are left with no clear procedure to measure, avoid, or fix such problems.

The data scientists building such models are also often in the dark, and may be unintentionally creating an unfair model. How, then, can an ethical data scientist do better? We will show that there are no simple solutions. For example, one can exclude race and other protected factors explicitly, but as one's data dimensionality grows, one will likely not be able to exclude them implicitly, through correlated attributes. This problem demands robust tools.

To introduce the set of available tools, we will first discuss the root causes of discrimination with what we hope will be a useful taxonomy, and second we will offer a broad survey of discrimination measures and "discrimination aware" data mining methods, with practical guidance. Although there exists several surveys on the topics discussed, we present them with the practicing data scientist in mind, as opposed to the research community. While the goals of these two communities are often aligned, sufficient differences in the definition of 'optimality' exist to warrant a different, more industry-focused perspective. Our intent is to spur data scientists to action, and for them to accurately report discrimination metrics to stakeholders & decision makers.

The rest of this article is organized as follows. In section 2 we will explore and survey common notions of discrimination[6], and provide quantitative examples. In section 3 we present a framework for identifying and auditing how some form of discrimination may enter into the data mining process; we hope that by addressing how certain commonly used tricks and heuristics may induce discrimination, practitioners can train themselves to practice caution when making their design decisions to avoid unintended effects. Section 4 describes a model development process updated to include discrimination awareness, and presents an overview of how different methods fit into this process, as well as practical advice to consider when choosing a method. Finally, in section 5 we analyze two case studies in which potential disparate impact may arise using our taxonomy and auditing framework.

## 2. What is Discrimination?

A core step in proactively reducing unintentional discrimination in machine learning systems is to understand what is discrimination in the first place. Coupled with both the conceptual and legal definition of discrimination is the means or metric to actually measure it within data or within a machine learning system. There is a vast literature on societal discrimination (see Altman SEP for an overview). To narrow the scope, we will focus primarily on classification and ranking systems, but the context of societal discrimination is squarely within view.

To narrow the scope of our discussion, we will focus primarily on classification and ranking systems.

---

[6] Unless otherwise noted, we'll use the word 'discrimination' to mean the "differentiated (and usually negative) treatment of an individual based on membership in some legally protected class," and not the more neutral form of "differentiate individual instances based on observed characteristics," which is generally the goal of machine learning.

## 2.1 Discrimination Starts with the Law.

There is no single way to define discrimination. In a given context, often the best we can do is to find a relevant and applicable legal notion of discrimination, and propose an associated "best fit" metric. We'll start by reviewing the social and legal concepts of discrimination as presented by Dabady et al,[4] which contains two examples of discrimination:

(1) Differential treatment on the basis of membership in a protected class that disadvantages members of that class (we'll call this 'disparate treatment')

(2) Treatment on the basis of inadequately justified factors, not including membership in the protected class, that disadvantages members of the protected class (we'll call this 'disparate impact')

The first case can be considered one of direct discrimination, where the protected class membership is directly considered in some decision. The second case arises when a protected group receives a different distribution (and usually more negative) of outcomes, without their protected group membership explicitly taken into account. This could happen unintentionally when making decisions on factors that happen to be correlated with membership in a protected class.

An example of disparate treatment is not giving a loan to someone who is African American, independent of any other personal characteristics. The explicit reason for rejecting the person, in this case, was the person's race. "Redlining" (refusing opportunities to people base solely on their zip code) for loans is a classic example of disparate impact. One's zip code likely has no material impact on credit worthiness, other than to serve as an implicit coding for some other, and potentially sensitive, variable. We refer readers to the discussion by Barocas and Selbst[1] for a review of legal frameworks for establishing liability for both disparate treatment and disparate impact. In later sections, we will provide a thorough taxonomy of the different ways either form of discrimination may find its way into a classification system.

## 2.2 Causality and Disparate Treatment

The literal language of discrimination law gives meaningful instruction on how to measure and detect disparate treatment in a decision process (both driven by machine learning systems or humans). Title VII of the Civil Rights Act makes it unlawful "to fail or refuse to hire … because of such individual's race, color, religion, sex, or national origin."

The courts typically require plaintiffs to make a direct causal connection between a decision in question and the affected person's membership in a protected class. This convention gives practitioners a well-established framework for the identification of discrimination in a decision process, namely the Rubin Causal Model using the framework of "potential outcomes[5]", which we will present after establishing some nomenclature for the rest of this article.

Let $S$ and $X$ be features of a person, where $S$ represents membership in some protected class (race, gender, religion, etc.), and $X$ is a vector of non-protected attributes, which is further divided into observable and unobservable attributes, denoted $X^o$ and $X^u$. In general, we'll assume the existence of some binary outcome variable of interest $Y$ in the training set and a decision process $M(S, X)$ that maps the inputs $X$ and $S$ to an action A. For simplicity, we'll assume that both $S$ and $A$ are binary, and use a superscript to denote realized values (i.e., $S^1$ for $S = 1$).

Here's a concrete example. X could denote attributes of past job applicants, such as experience, education, and zipcode, which we can assume are all observable; S could be "1" for minority and "0" for white applicants; and Y could be "1" if the person was offered the job and subsequently stayed at the company for at least 2 years. The decision process *M(S, X)* would train on past data with target Y, and map a new applicant to a probability of being offered a job and staying for at least two years. If that probability is found to be above some threshold, the action A could be defined as "1", which is to say a job offer.

The potential outcomes framework imagines a counterfactual, parallel world, where we can observe the value of an outcome variable after changing only the value of one variable of interest. In our case, we want to observe the outcome *A* (whether the person is offered the job) on any individual by changing only $S$ from $S^1$ to $S^0$ (the race of the candidate). We can establish a causal link, and thus deduce disparate treatment, if in our counterfactual framework we can observe that, all else being equal, a change in the value of *S* changes the outcome. For any arbitrary decision process we would state this formally as:

$$\text{Causal Mean Difference} = E[A|S^1, X^o = x^o, X^u = x^u] - E[A|S^0, X^o = x^o, X^u = x^u]$$

And note that this is different from the following expression:

$$E[Y|S^1, X^o = x^o, X^u = x^u] - E[Y|S^0, X^o = x^o, X^u = x^u]$$

The first expression establishes whether the expected outcome of a decision process is causally linked to a protected attribute. The second establishes whether the target variable *Y observed in the data* is causally linked to the protected attribute. There is of course a relationship, because if the training data has problematic features, one could easily imagine it informing the end process.

However, establishing the causal link using training data is a difficult and controversial process without the use of randomized experimentation, simply because you might not have many examples of people sharing all attributes except S. In our example, for example, you might not have many examples of two candidates with the same zipcode, education, and job experience but different minority status.

On the other hand, establishing causality in the decision process is rather trivial, at least when there are no unobservable attributes[7]. Given a classification system under audit, all one needs to do to establish its disparate treatment is to evaluate a set of representative examples (possibly synthetic), and average differences in output when all are set with $S^1$ and $S^0$. This could be further refined for any partition of *X*.

The implication is strong here for data scientists. Including a protected attribute in a classification system runs a tremendous risk of liability under disparate treatment doctrine. In certain domains, such as employment and credit, many of what we often refer to as 'protected' attributes are formally outlawed. However, outside of explicitly formal laws governing the use of such attributes, data scientists should be aware that, even with the best of intentions, using commonly protected attributes exposes them to potential legal risk.

In later sections where we survey various discrimination-aware data mining techniques, all solutions require access to these protected attributes. This presents an inherent challenge for data scientists. Lack of knowledge of a protected attribute may mitigate the risk of unintentional disparate treatment (but not disparate impact as we will see), but it also reduces one's ability to detect and avoid it. This is a paradox that will require careful negotiation between firms, legal oversight and society to resolve.

---

[7] This is often untrue, especially if managers are interviewing candidates in person for example, and get to score the candidate. It's more true if there are practices in place to extract information without the possibility of bias, for example in orchestra "blind auditions."

## 2.3 Observational Measures and Disparate Impact

As discussed above, disparate treatment is a form of discrimination that may be more straightforward to both define and measure. Proving its existence may be a straightforward process of establishing a counterfactual around the decision apparatus, and demonstrating that the value of a protected attribute directly causes a different outcome. Establishing disparate impact may be less straightforward, and this is evidenced by the fact that multiple competing approaches have been established in the research literature. In this section we'll review different approaches for measuring disparate impact, point the reader to resources for further study, and connect these approaches to the causal framework we discussed above.

Zliobaite provides a comprehensive survey on metrics for measuring disparate impact[6], which we will summarize here. The earliest established measures are categorized as *statistical measures*, which are formal procedures to test the validity of a statistical hypothesis. These typically indicate the binary presence of discrimination, but do not necessarily measure its magnitude. A straightforward example is based on regression methods. In this approach a model is specified, such as $A = \alpha + \beta*X + \varphi*S + \varepsilon$, and the sign and statistical significance of $\varphi$ is taken to establish the presence of discrimination. In a recent paper studying racial bias in police shootings, Fryer et al. use such a regression based approach[7]. Further examples can be found in the survey paper by Romei et al[8]. Much like with statistical tests in any application, statistical significance does not necessarily mean practical significance. Minute levels of discrimination may be statistically significant, though within a socially or legally tolerable range.

The machine learning literature on discrimination-aware data mining tends more to employ what Zliobaite refers to as *absolute and conditional measures,* where the latter can be viewed as an extension of the former. The most general form of an absolute measure is structurally similar to the counterfactual presented above, though because it does not directly condition on *X*, is more of a correlative than a causal approach:

$$\textit{Mean Difference} = E[A^1|\ S^1] - E[A^1|\ S^0]$$

This score, or an equivalent variation, is used by several authors, despite being associated with different discrimination-aware algorithms[9,10,11]. Zliobaite offers, and recommends with empirical analysis, a normalized modification of the *mean difference*, dividing it by the following normalization constant[12]:

$$C = \min(P(A^1) / P(S^0),\ P(A^0) / P(S^1))$$

Normalizing the mean difference this way produces a score of 1 or 0 at the maximum and minimum levels of discrimination.

One dominant issue with absolute measures is that the different protected populations of interest may be distributionally different over the domain of $X$, and that any discrimination observed at the macro level can be explained away by other, non-protected attributes. We can control for the explainable portion of the group level discrimination by modifying the mean difference to condition on observed, non-protected attributes:

$$\textit{Conditional Mean Difference} = E[A^1 | S^1, X^o = x^o] - E[A^1 | S^0, X^o = x^o]$$

This modification looks for whether expected outcomes across the different values of the protected attribute differ after controlling for observed, non-protected characteristics.

As an example, consider credit approval discrepancies between men and women. At the population level, we may see differences, but after controlling for attributes such as income and credit score, we may see no difference.

Kamiron et al incorporate this notion into a metric called *unexplained difference*[13], which loops through the entire data population attempts to decompose total mean difference into explainable and unexplained portions.

## 2.4 Further Connections to Causal Inference

The connection to causal analysis has not been lost on the discrimination research community, and several measurement techniques draw directly from this body of work. A common problem when conditioning an expectation on a particular value $X = x$, is low support at that exact value. Many researchers in the causal literature remedy this issue by grouping observations with similar values of $X$ into a cohort and taking differences within the cohort. Luong et al show that this approach can be done for discrimination measurement using k-Nearest Neighbors to match on $X$[14]. Calders et al similarly use "propensity score stratification" to group observations into cohorts, where the propensity score is based on the likelihood to be a member of a protected class[15].

Note the similarity between the conditional mean difference and the causal mean difference defined earlier. The main, and notable, difference is that the causal difference is also conditional on unobserved characteristics.

A core assumption in causal inference is that are no unobserved confounders. This assumption is usually met in randomized controlled experiments, but is generally difficult to prove in observational studies. The same assumptions and implications exist in measuring discrimination. Whether an outcome is caused or simply just correlated with a protected attribute can be a deciding factor in establishing disparate treatment or disparate impact.

As mentioned above, when auditing a classification system that takes in a fixed input set of features *(X, S)*, there are no unobserved attributes that would affect the outcome. Thus, the conditional and causal mean differences are equivalent. When auditing data (either a 3rd party auditor auditing a classification process with incomplete knowledge of the input set, or a data scientist auditing their training data), we can not assume there are no unobserved attributes that affect the outcome. In such cases, claims for disparate treatment should be much harder to satisfy, relying as they do on technical definitions of causality.

## 3. Where Good Algorithms Go Bad

It is easy to dismiss claims that algorithmic decision-making produces unintentional discrimination by standing behind the supposed "objectivity" of data driven decision-making[8]. As of writing this article, the research community for discrimination-aware data mining is relatively small, and the set of published, and most importantly, commercially validated research is likewise limited. Prior efforts have been made to cover ways in which discrimination can emerge in a classification system[1, 16]. In this section we add to and synthesize the prior efforts, and we organize the discrimination drivers into a taxonomy.

Our goal is twofold: first, to educate and convince data scientists that, even with the best of intentions, some form of discrimination may arise in their systems, and second, to connect these causes to stages within the model development lifecycle so that the practicing data scientist can both anticipate and proactively manage these potential problems.

---

[8] In casual conversations with colleagues and other data scientists, in person or via social media, the authors have encountered such debates and thus believe that there is some degree of running skepticism amongst data scientists over algorithmic discrimination.

3.1 **Model Building 101**

Figure 1 shows a commonly accepted process diagram for model building and deployment. This is known as the Cross Industry Standard Process for Data Mining (CRISP-DM)[17], and has been a standard for well over a decade.

This diagram shows how data science tasks (particularly deployed supervised learning systems) are commonly approached as an iterative process, where the feedback from prior iterations informs models and decisions in subsequent steps. While the heart of a classification system is the training step, it is generally well understood by data scientists that training the actual algorithm comprises the minority of a data scientist's time. The majority of time spent is generally focused on making somewhat subjective decisions, such as what events to predict, where and how to sample data, how to clean said data, how to evaluate the model and how to create a decision policy from the algorithm's output.

Our taxonomy will show that discrimination can creep in at any one of these stages, so persistent vigilance and awareness is advised throughout the process. When we discuss solutions later, we will also make recommendations on where in this process a data scientist can employ discrimination-aware unit tests to appropriately audit the need and effectiveness of their chosen discrimination-aware techniques.

*Figure 1:* Process diagram for the CRISP-DM model. This model generalizes the commonly used practice for building and deploying a machine learning model.

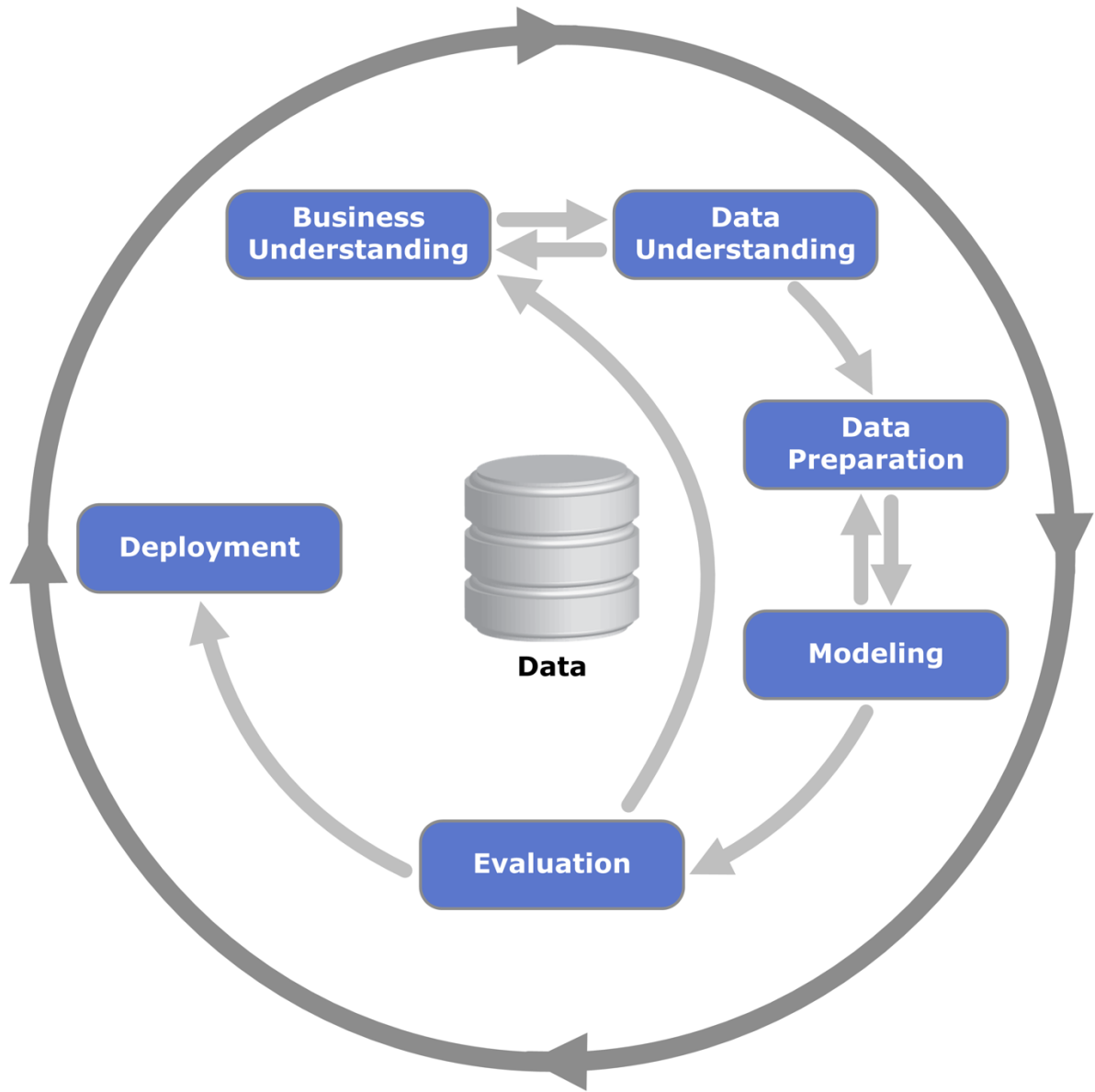

### 3.2 A Taxonomy of Causes

A classifier alone does not discriminate, but unintended discrimination can find its way into a classification system in various ways, and at all parts of the data mining process.

We focus on building and on using the classifier. For each procedural step of the building process, we categorize the discrimination by cause: either data issues or misspecification. When we consider how the classifier is used, we include in our taxonomy "procedural failures" that could lead to discrimination.

Figure 2 shows a high level view of the taxonomy, along with indicators showing in what parts of CRISP-DM we might expect to encounter each cause. After providing more detail on each entry in the taxonomy, we will follow with a survey of remedies for the practicing data scientist to consider when faced with these issues.

Figure 2: Commonly cited causes of classifier discrimination.

| Component | Root Cause | Level 1 Description | Level 2 Description |
|---|---|---|---|
| Classifier (Model Development) | Misspecification | Target Variable | Hetergeneous Target Variable (A,B,C,D) |
| | | | Proxy Target Variable Learning (A,B,C,D) |
| | | | Target Variable Subjectivity (A) |
| | | Model/Feature | Inclusion of Protected Attributes without Control Variables (C,D) |
| | | | Inclusion of Protected Attribute Proxies (Redlining) (C,D) |
| | | Loss Function | Omitted Discrimination Penalties (D) |
| | | | Failure to Specify Assymmetric Error Costs (D,E) |
| | Data Issue | Sample Bias | Discrimination In Data (A,B,C,D) |
| | | | Under-representation of Protected Class (B,C,D) |
| | | | Over-representation of Protected Class (B,C,D) |
| | | | Low Support (B,C,D) |
| Classification System (Process Failures) | Failure to Audit | | Data Preparation (B,C) |
| | | | Classifier: Pre-Deployment(D,E,F) |
| | | | Poor Feedback Loop (E,F) |
| | | | Non-Deferrence to Human Experts (A,E,F) |

Caption{Letters detail where in the CRISP-DM process a data scientist is likely to encounter each cause, corresponding to: A = Business Understanding, B = Data Understanding, C = Data Preparation, D = Modeling, E = Evaluation, F = Deployment.}

### 3.2.1 A Classifier's Source of Discrimination

For this taxonomy we separate the classifier from the classification system. We define the classifier as the function that maps an input tuple *(S, X)* into an action space *A* (i.e., hire/don't hire, assign a police patrol or not, etc.), and the classification system as the technology and set of processes that implement the classifier.

#### 3.2.1.1 Data Issues

The most straightforward cause of discrimination can be summed up by repurposing a commonly used phrase in computer science: discrimination-in, discrimination-out. If we assume a perfect development scenario of unlimited data, or big enough to reach asymptotic estimation limits, and sufficient computation resources, we can learn a classifier that reaches peak theoretical performance guarantees. This is called a Bayes Optimal Classifier; such a classifier would be a perfect mirror of the data generating distribution that trained it.

What if such a distribution itself was inherently discriminatory? If we think of our society as a data generating process, and we acknowledge that, despite great progress over the years, our data as a cultural echo still reflects systemic biases against protected classes, then we should not be surprised to find discrimination in our models. This was expressed quite eloquently by Pedreschi et al.[18] in a prescient paper written nearly a decade ago:

> *"[L]earning from historical data may mean to discover traditional prejudices that are endemic in reality, and to assign to such practices the status of general rules, maybe unconsciously, as these rules can be deeply hidden within a classifier."*

Another type of bias, sample bias, can result in a discriminating classifier, even if the underlying data generation was free of protected class bias. Our taxonomy shows two flavors of sampling bias: over and under representation. Barocas et el give examples of the risk of overrepresentation in the employment context, noting that managers may give disproportionate attention to a protected class group, and the increased scrutiny may lead to a higher probability of observing a target transgression[1]. A specific example could be managers who monitor ex-convicts, or African Americans due to some preconceived prejudice. In a world where some negative behavior (like tardiness) is not correlated with a protected class, disproportionate sampling (monitoring) of a given class may increase the likelihood of observing the negative behavior and thus induce a statistical correlation in the data.

On the other end of the sampling bias spectrum is underrepresentation. We also include 'low support' in this discussion, as although we have grouped them separately in our taxonomy, both suffer from similar statistical issues. By 'low support', we mean that a protected group $S^0$, has a smaller sample size relative to other groups in the data. This is compounded when we think about logical conjunctions of protected groups (African American women vs women or African Americans). Differences in sample sizes between groups mean that more represented groups will dominate the learning algorithm. Kamishima et al call this phenomenon *underestimation*[19], and characterize it as a learning algorithm failing to converge on a value of *X* that has low support in the data, and thus reverting to average trends in the data to avoid overfitting. Underestimation is even more likely to happen in cases where the target variable is rare (credit default, violent crime, etc.) and pruning or regularization methods are employed.

The extreme of low support is a complete lack of representation in the data. In this case, the model suffers from an *identifiability* issue, which is that one cannot model outcomes on individuals never observed. Whatever the cause, low support of a group can lead to total exclusion of this group in future positive actions made by the decision process, as the classifier may not learn about any positive behavior that characterizes the group.

### 3.2.1.2 Misspecification

A common concept in statistics is the idea of model misspecification, which can be described as the functional form or feature set of a model under study not being reflective of the true model. For example, if you try to impose a linear model on a variable that depends quadratically on the inputs, that would constitute a functional misspecification. We extend the idea of model misspecification to misspecifications of the underlying problem – for example in the form of an imperfect proxy – as well as misspecification of the loss function used to train the classifier, which might not assign a sufficient "cost" to a false positive or false negative.

In the age of highly complex classification algorithms such as Random Forests and Neural Networks, functional misspecification isn't usually a concern since data scientists can rely on complex, non-linear algorithms to approximate any function, at least given enough data. But a model is more than its functional form, and most anti-discrimination efforts will lie in the other choices of model specification.

Let's start with inputs. Including the protected attribute $S$ might be useful in a predictive sense, but of course this puts the system at a much higher legal risk. For that matter, $S$ itself might not be a cause of the underlying discrimination, but might be correlated through the omission of other, non-protected attributes. For example, women may be observed to have higher default rates than men, but after controlling for income, this correlation may disappear or even be reversed. Using a protected attribute as a proxy for other variables that truly carry the signal could easily result in a discriminating classifier.

In the converse situation, where the modeler omits $S$ but leaves in non-protected attributes $X$ that are both correlated with $S$ and the target variable $Y$, we may end up with a model that exhibits the so-called *redlining* effect. This could be a likelier scenario than the one presented just above, as many domains explicitly prohibit the use of *S* but not variables that carry the signal of *S*. As this is a fairly common scenario, many of the research methods we cover are explicitly motivated by this phenomenon.

Next, we identify several forms of misspecification of the target variable. These all stem from the fact that a data scientist often has several choices on how to define a target variable, and the decision process itself can be subjective. Barocas and Selbst[1] give a good example of this for the case of building classifiers to predict a future "good" employee from a "bad" one. The target itself is a subjective one, and could include any part from a set of factors such as tardiness, performance review, tenure, etc. In such a case one may be able to find systematic correlations between a protected attribute and a subset of these factors. The final model's discrimination could then depend on the subjective choice of how the target was defined.

For example, tenure at a job might be correlated to gender, but this correlation might be the result of a workplace that is more pleasant for one gender of another. Importantly, it might be uncorrelated to whether the given employee was actually qualified in the first place, and replacing tenure with a different proxy for "good" could reduce the disparate impact.

Even if we assume that we can control for any apparent subjectivity, there still may be ambiguity. Certain problems deal with a heterogeneous set of events that can be naively labeled as a single type of outcome. An example is in crime prediction for predictive policing. Although there are many types of crimes, we can simplify the example by considering violent felonies (rape, murder, armed robbery, etc.) and petty misdemeanors (loitering, low-level drug possession, shoplifting etc.).

A common objective for communities would likely be to reduce violent crime above all else. A data scientist facing this predictive task may frequently find, after a quick analysis of the data, that data on the primary objective is sparse. For statistical reasons, the model's predictive ability seems to increase if the target is supplemented with lesser crimes. One approach could be to mix all types of crime events into a single label. Another, called *proxy learning*[20], uses an alternative proxy label that is correlated with the main label of interest, but is much more abundant in the data. In either case, the event that is most frequently observed in the data will dominate the direction the model takes, as the model weights will be adjusted more frequently by the penalties on that particular event.

Critically, if the original event of interest (say violent crimes) had little to no correlation with a protected attribute, but the additional events mixed into the label do, the final classifier should be expected to discriminate against that particular attribute.

Lastly in our discussion of misspecifications, we consider cost function misspecification.

An important example is a failure to consider different misclassification costs. This omission is directly connected with our discussion of heterogeneous target variables. In the crime prediction example, one way to give equal or more weight to the primary event of interest – violent crime – is to give those examples, or the errors induced by them, more weight in the training function. This is called upsampling or importance weighting[21].

More generally, data scientists should always be sensitive to misclassification costs, since the act of discrimination is usually enforced when a protected class received an unwarranted negative action. If a false positive error can have significant harm to an individual in a protected class (such as the case of not hiring someone, or using a predictive model to determine a convict's parole), the classifier and classification system should carefully take into consideration the asymmetry of the misclassification costs[9]. Straightforward methods outside of the discrimination literature have been proposed to address cost sensitive classification[22,23], and these should be part of the data scientist's toolkit for conscientious classification.

Another form of cost function misspecification is an error by omission, which we define as leaving out or ignoring terms that penalize for discrimination. This idea is more novel, and many standard learning packages don't accommodate this. Zemel et al. and Kamishima et al. both introduce augmented cost functions, where they augment a standard log-likelihood loss function with a 'fairness' regularizer[11,19]. Fairness here is an additional concept that measures how well people who are similar to each other receive similar treatments or predictions. This additional constraint ensures that discrimination awareness doesn't lead to an adverse level of unintended affirmative action.

The fairness aware regularizers take into account differences in how the learning algorithm classifies protected vs. non-protected classes (i.e. $S^1$ vs $S^0$) and penalizes the total loss based on the extent of the difference. Much like in standard L1/L2 regularization, the penalization is controlled by a tunable hyper-parameter, chosen via some form of cross-validation. Note that in this case, the final objective function would have an accuracy component and a discrimination component. A data scientist can tune the hyper-parameters to minimize the global loss function, or can minimize one component subject to a constraint on the other (i.e., minimize discrimination up to a maximum misclassification rate).

---

[9] The logic here actually applies to all classification systems, not just those exhibiting disparate impact or treatment.

### 3.2.2 Process Failures in Developing the Classification System

A classifier does not itself discriminate. It takes a human process that includes that classifier to discriminate. Therefore a data scientist needs to remain vigilant throughout the entire evaluation and deployment process. The greatest error one can make during evaluation is to not test an algorithm for potential discriminatory behavior. This is an error that starts with the data scientist but should extend upwards to the executive team. As the developer and steward of the algorithm, the data scientist should meet with all constituents to understand what are the (potentially competing) objectives. Teams should discuss and agree upon a set of legal and ethical standards for each type of problem. With a concise set of standards for what discrimination means to the application, and what are the allowable boundaries, the data scientist can select an appropriate discrimination metric and begin a series of discrimination unit tests on decisions made by the system.

Another process failure involves designing appropriate feedback loops. If a system's predictions can produce individual harm, it should be the responsibility of internal managers to monitor and verify the validity of the system's output. Additionally, when designing the classification system's feedback loop, the data scientist must account for the distortion that any model-based intervention may have on the data generating or sampling distribution.

An example of this arises in predictive policing, where non-violent crimes are more likely to occur (or really, be observed and registered) where police officers patrol. As most systems are trained on production data, once a model goes live, the distribution of the data the production system collects will shift towards the decisions made by the system. Such distortions can create extremely negative vicious cycles, where the interventions induce statistical deviations between protected and non-protected classes which then get learned and perpetuated by the next iteration of the algorithm.

Finally, while data scientists should approach their work with scientific rigor in mind, we encourage an active and healthy level of skepticism towards all models and all results. Part of this skepticism should include deference to human experts where applicable. One seldom finds a classifier with perfect or even near-perfect predictions. There will always be hard-to-classify points near a decision boundary, or irreducible error in the target variable, and in such cases, the type of systems we have been discussing here can do possibly irreparable harm to an individual or specific groups.

Part of being a responsible data scientist is knowing how and when to involve human experts in the decision making process. *Active Learning* is a branch of machine learning that focuses on cost efficient ways to engage experts in a way to improve future iterations of a model[24], and this body of work can provide tools for data scientists to not only address discrimination, but to mitigate classification errors that have a high human cost.

Additionally, classification systems are limited by the data at their disposal. A parole board making a judgment solely based on predictions derived from a classifier with a static set of features (such as criminal record and responses from a survey) may miss an opportunity to incorporate recent experiences with the convict, or any information outside of the observational scope of the classifier. In general, we should never expect any classifier to be feature complete. Keeping humans in the loop improves objectivity and nuanced flexibility.

While we do not claim that the above set of issues and rules is exhaustive, we feel it is a comprehensive starting point for the practicing data scientist to learn to become aware of how unintended discrimination may occur. Awareness is a necessary but not sufficient condition for preventing discrimination in classification systems.

## 4. **Auditing and Removing Unintended Discrimination**

In this section we propose a discrimination-aware auditing process that mirrors the standard process for developing classifiers. It has been amended to incorporate discrimination unit tests and highlights where state-of-the-art remedies may fit. We also present an overview of prior discrimination-aware data mining work as a reference to readers, with suggestions for how a data scientist might think about incorporating different methods into their common workflows. By developing discrimination-aware systems, data scientists can incorporate ethical and legal constraints into their models, and thereby result in the intended outcomes that ethics and law hope to achieve.

Figure 3

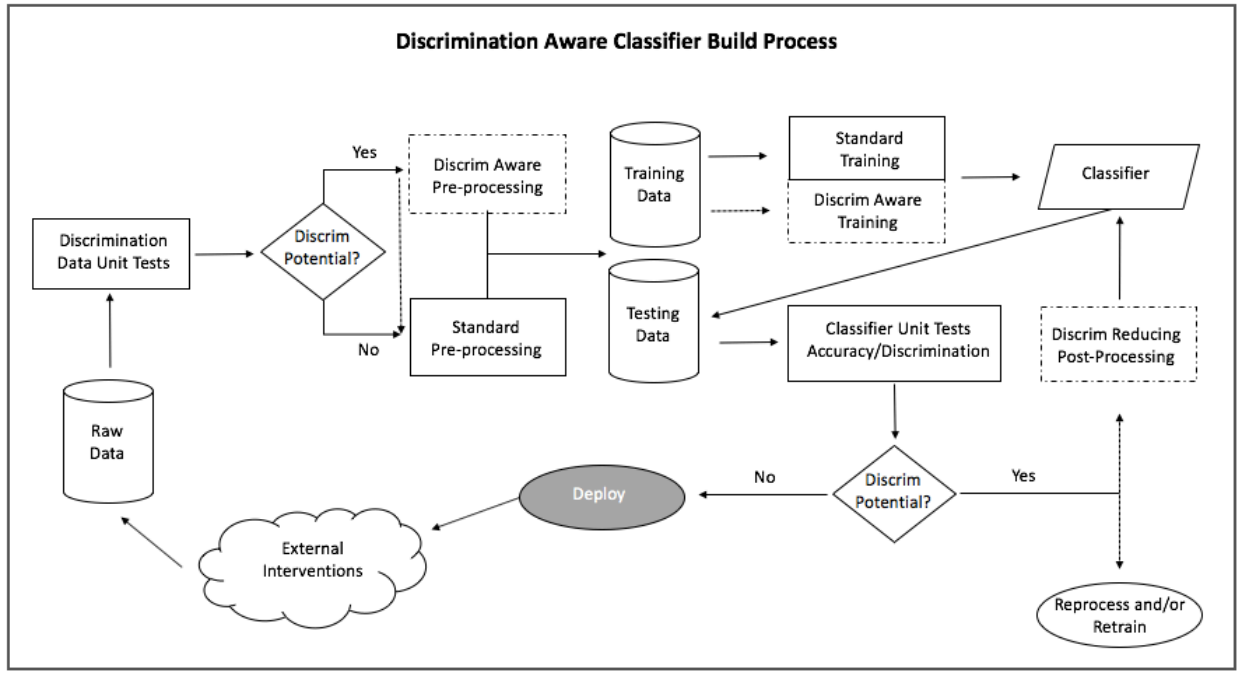

Caption {We update standard a classification process diagram to reflect key testing and decision points that can inject discrimination awareness into the workflow. Dotted arrows and figures represent optional workflows, representing pre-processing, in-processing, and post-processing techniques, as suggested by the prior art}.

Our process in Figure 3 begins with the raw data to be used for model development. We will assume here that the scientist has already carefully formulated the business problem and has consulted with the appropriate managers and legal team to understand the applicable anti-discrimination laws and/or ethical standards governing the application. This prerequisite work should define what data is available as a target variable and as predictive features, and what the appropriate discrimination metric should be.

Given this set up, we propose several "discrimination aware" unit tests that may guide the future development of the classifier. These tests fit into the process where exploratory analysis usually sits.

The first test is to use the chosen discrimination metric to determine if the underlying data discriminates against a protected attribute with respect to the target variable $Y$. These tests should include any separately measureable event that may be rolled up into the target variable (see above discussion on heterogeneous target variables for motivation). Next, the data scientist should test and document any features that have medium to high correlation with the protected attribute. Finally, for each level of each protected attribute, and possibly conjunctions of multiple attributes, the support ($P(S^i)$), which is to say the proportion in the overall population, should be computed to learn about which protected group errors may be more susceptible to statistical estimation errors.

When any of the above tests score above or below a certain threshold (which should be set by the applicable legal and ethical standards), the data scientist can remove the discrimination via methods that fit into multiple stages of the process. For more comprehensive coverage of specific methods, we direct readers to prior surveys of the field[8,16]. We present in this section several high level methods following the organization nomenclature of Hajian et al from their tutorial in the 2016 ACMSigKDD conference.

The discrimination-aware methods are grouped into *pre-processing*, *in-processing*, and *post-processing* techniques. These fit into our diagram in Figure 3 as techniques that can be executed in the data pre-processing, training, or post-scoring evaluation steps (respectively). Figure 3 shows these in diagrams with dotted lines, indicating that the data scientist may choose one or a combination of these techniques.

*Pre-processing techniques* attack the problem by removing the underlying discrimination from the data prior to any modeling. Kamiran and Calders present three promising pre-processing methods that are competitive to other non-pre-processing techniques[25]. These methods involve one of strategically relabeling examples near the classification margin, reweighting or resampling examples, and they particularly endorse the relabeling or resampling techniques. These are extended in future work to account for cases of conditional, or explainable discrimination. Feldman et al present another method which removes dependencies between *X* and *S* but still allows *X* to be predictive of *Y*. They acknowledge the tradeoff between fairness and accuracy and design the method so that the tradeoff can be tunable and tailored to a specific application. There is no direct comparison between these two methods, though we see more implementation barriers in the Feldman approach[26]. Kamiran et al present clear instruction on their methods with minimal mathematical notation, making the learning curve for adoption potentially shorter for data scientists looking for a quick prototype. Additionally, Feldman et al admit limitations on the applicability to non-numeric data (i.e., categorical), whereas the methods by Kamiran et al appear to be operable on all data types.

*In-processing* techniques can be thought of as modifications of traditional learning algorithms to address discrimination during the model training phase. This is referred to as 'Discrim Aware Training' in the process shown in Figure 3. Zemel et al. and Kamishima et al. both introduce augmented cost functions, where they augment a standard log-likelihood loss function with a 'fairness' regularizer[11,19]. These regularizers take into account differences in how the learning algorithm classifies protected vs. non-protected classes (i.e. $S^1$ vs $S^0$) and penalizes the total loss based on the extent of the difference. Much like in standard L1/L2 regularization, the penalization is controlled by a tunable hyper-parameter. Zemel et al give examples on ways to tune hyper-parameters to achieve different outcomes. Other *in-processing* techniques involve discrimination specific modifications of traditional learning algorithms, such as Naïve Bayes and Decision Trees[9,10].

*Post-processing* is the final class of methods and can be performed post-training. It relies upon access to a holdout set that was not involved in the model's training phase. We highlight a recent contribution to the prior art and focus on some highly desirable properties of this class of approaches. Hardt et al propose a post-processing technique that works very well in our present environment of black box classification systems. The elegance of this approach is that it relies solely on access to the prediction (*A*), protected attribute (*S*) and label (*Y*) in the data, and can be executed with only aggregated data. Details between the mapping of *X* to *A* are not needed, making this approach very amenable to situations in which third parties (either internally or externally) are engaged to independently de-discriminate the classification system. Despite these promising properties, there is no empirical evaluation between this approach and others, and the exact implementation details are somewhat obscured by mathematical detail.

When and how to choose both a discrimination measure and a discrimination reducing method are critical questions that we acknowledge are perhaps too nuanced and complicated to offer a single blanket approach. The research literature provides inconsistent guidance on which methods to choose. They generally evaluate each method on two dimensions: accuracy vs. discrimination. Moreover, empirical comparisons are difficult to make because different researchers use different measures of discrimination depending on the study.

Additionally, while maximizing accuracy given a legal threshold of discrimination is an appropriate objective for research, many data scientists and firms do not operate with such simplistic goals. Hand laments that typical research comparing classifier performance fails to consider the complete set of constraints or objectives that firms consider when evaluating options[28].

As practicing data scientists ourselves, we can testify that firms tend to prioritize ease of implementation, run time scalability, and low risk of failure as high as model performance. Data science managers often advocate for using straightforward modeling techniques and focus on execution speed[29]. This is in spite of the fact that, as can be seen in the empirical studies we cite[11,16,26], differences between the highest performing state of the art and previously established methods are often within a few percentage points of each other.

With all methods performing within a narrow competitive range, we would suggest that data scientists should give preference to those that can be easily adapted to existing discrimination measurement processes or that can still leverage commonly used open source training libraries (such as pre & post-processing).

Discrimination awareness starts with the awareness level of the data scientist herself, and learning the taxonomy of causes as laid out in this paper is real step closer to maintaining that awareness in all aspects of the data mining process. From here, the data scientist should seek out and possibly provoke or even lead the discussion with legal, ethical and management experts to translate the governing standards into an appropriate quantitative and qualitative definition of discrimination. From there it is a matter of execution, and we have offered a process to guide the data scientist, as well as a brief survey of methods as a starting point to consider what is right for the firm and the application.

## 5. Case Studies

In lieu of pure empirical results (which would normally occupy this section of an article), we offer a few case studies to highlight both the highly dependent nature between problem specifics and optimal design, as well as how the general auditing framework we presented above can be applied to different problems[10]. While there are numerous predictive modeling applications to choose from for case studies, we chose those that tend to have high social and human costs while being fairly opaque.

---

[10] These case studies have been adapted from the research and discussion performed by one of this paper's co-authors, in the book *Weapons of Math Destruction*. Unless cited otherwise, references for the case studies can be found in the book[2]

## 5.1 Predictive Policing (PredPol)

*Synopsis*

Facing economic pressures, towns across America are using predictive policing to more efficiently address crime. Example companies providing such predictions are PredPol, Compstat and Hunchlab. The general intervention framework is to predict geographic zones in which crime is most likely to occur and station more police surveillance in the riskiest zones. These crime predictions typically don't classify individuals, and use historical crime counts by time for each location as primary features. Predicted crimes generally are categorized as either 'violent' (rape, robbery, homicide, etc.) and 'nuisance' (loitering, panhandling, possession of small quantities of drugs) crimes.

*Audit*

Our first observation is that we see opportunities for predictive policing to suffer from multiple forms of target variable misspecification. Overall policing strategy is often public record so any police force using predictive methods should be clear and precise about its problem statement. 'Making neighborhoods safe' is an example of an ineffective and imprecise problem statement (for predictive modeling, not political maneuvering). A better example would be 'Reduce the incidence of violent crime without increasing the police force.' Given this objective, past crimes labeled as such would be enumerated and used as a dependent variable in some supervised learning algorithm. Violent crimes are likely to be rare, which is compounded by the need to predict at the level of small geographic zones in a short time interval. A natural solution to this sparsity problem would be to use nuisance crimes as a proxy variable for violent crimes.

However, nuisance crimes are crimes that disproportionately impact poorer and minority clustered neighborhoods. These crimes are often symptoms of broader social issues, including poverty, homelessness and unemployment. One issue with using these as proxies is the lack of support showing sufficiently high correlation between nuisance and violent crimes. The assumption that there is a high correlation is exemplified by 'Broken-Window' or 'Zero-Tolerance' policing initiatives, and these are highly controversial due to disparate impact and researched claims of lack of validity.

In this setup we identify either a heterogeneous set or a proxy used as the target variable, and the dominant event is one that is known to be correlated with protected attributes, with dubious correlation to the primary event of interest. Such qualitative likelihood of disparate impact should warrant particular scrutiny using an auditing process such as that discussed in this article.

If a police routing process showing disparate impact were deployed, the nature of the feedback loop could perpetuate the disparate impact in future generations of the model. This is particularly true where nuisance crimes were part of any human or automated decision function. Interventions recommended by the system could significantly alter the sampling distribution, in that nuisance crimes are only recorded if an officer were there to observe it. This differs from many cases of violent crimes, where officers are generally called to the crime scene.

In a pre-predictive world, if police forces used any form of racial profiling (which has been highly documented in New York City's controversial stop-and-frisk program), then we would expect to see more nuisance crimes in racially profiled neighborhoods. Fryer et al. have used regression techniques to prove that such systematic police bias is directed at protected minority groups[7]. Any predictive model derived from this data would likely reinforce these patterns, as nuisance crimes likely just follow where people tend to cluster, and are thus much easier to predict than violent crimes. If the predicted crimes are disproportionately in poorer minority neighborhoods, the biased sampling and lack of correction would likely start a terrible vicious cycle.

As data scientists, we see statistical validity in mixing the nuisance proxy with the violent crime objective, at least where data shows some correlation between the two types of events. However, doing so without any form of corrective measure is an error that likely could steer the model towards disparate impact. Even if a predictive policing system offers more accountability (and hopefully less bias) than its non-predictive counterpart, we can not simply assume that systemic prejudice will just fade away upon its implementation.

To summarize this case's potential for socially destructive disparate impact, target variable misspecification coupled with a negatively reinforcing feedback loop could lead to a self-perpetuating system that continuously targets poorer and more minority concentrated communities. Fixing any broader issues of community-police relations may be outside of the scope of data science, but utilizing predictive policing fairly certainly is not.

This case also illustrates that the benefit of understanding root causes of algorithmically induced disparate impact is not limited to the data scientists or their managers. All participants of predictive policing policy stand to gain by developing some literacy around the problem, and should be made familiar with the applicable metrics, auditing methods, and mitigation techniques.

## 5.2 Kronos (Classifying Job Applicants)

*Synopsis*

Firms that manage large divisions of hourly-wage employees are increasingly turning to third party companies to administer screening tests to make assessments about an applicant's suitability for employment. One such company is Kronos, which was founded in the 1970's by MIT grads and has been a leading provider of workforce automation tools. These tests, for legal reasons, avoid questions that could offer direct evidence of inclusion in a protected class, and instead focus on a large set of personality questions. Example questions are: "It is difficult to be cheerful when there are many problems to take care of" OR "Sometimes, I need a push to get started on my work." The responses to these and similar other questions are used to assess an applicant's potential productivity and expected tenure. Applicants scoring at the bottom end are generally never hired and aren't given feedback about their assessment.

*Analysis*

Hiring is one of the most heavily governed decision processes in terms of anti-discrimination law. We acknowledge the value in standardizing what can easily be a very subjective process, subject to much personal prejudice. We also acknowledge that should a standardized and data driven system show disparate impact (disparate treatment is downright illegal), the purported 'objectivity' of the algorithm could perpetuate this impact and render a protected class vulnerable.

In this particular case we consider disparate impact against people with any degree of mental illness (diagnosed or not)[11]. Similar to the predictive policing case, we identify here a clear example of target value misspecification. Based on marketing material from the Kronos website, we deduce the following problem statement: "Use a predictive model to predict an applicant's productivity levels and likelihood to churn."

Churn, which refers to how quickly a worker needs to be replaced due to quitting or firing, is fairly cut-and-dry metric to observe. But productivity may not be as objective, nor will it necessarily be consistent across firms. Any subjectivity in both the definition and labeling of 'productivity' can cause statistical imbalances in the training data between the levels of protected class and the outcome.

---

[11] As of this writing, the practice discussed in this case is being challenged in the courts on the grounds that it violates the Americans with Disabilities Act by requiring a mental health exam as part of a hiring process.

All forms of target value misspecification that we discuss above may be applicable here; the exact form is dependent on the modeling choices made by Kronos data scientists. Given this potential for discrimination, the fact that discrimination in hiring decisions can impose significant costs, there should be no question that discrimination-aware auditing be included in this model development process.

We also see a potential for model misspecification here. Personality traits, measured via a candidate survey, are used here to determine employment fitness. Prior research has shown that personality traits are generally a poor predictor of workplace productivity[30], but they may be highly indicative of some degree of mental illness, which is a protected attribute in employment law. This specific form of model misspecification could result in a redlining effect against this particular group.

Our biggest concern here is the lack of an appropriate feedback mechanism. Considering the subjectivity of the target variable, prior research that invalidates the connection between workplace productivity and personality traits, and the potential for redlining, we should at least expect such a system to be highly accurate to justify the potential legal risks to the employer.

But the feedback loop is incomplete: we only observe the productivity outcome for applicants that are hired. This creates multiple problems. The first is that the original training dataset is likely biased and reflects the decisions of potentially biased people-based screening systems. Any previously affected protected group would be severely underrepresented (rendering this a *low support* data issue), and this could perpetuate disadvantages for this group. The other problem is that we can only observe false positives (assuming a 'positive' classification means 'hireable') and not false negatives.

In other words, we know how many bad candidates passed the exam (precision), but we do not know how many good candidates were overlooked (recall). This is an issue with respect to discrimination because if a particular protected class is excluded more often than others, there is no way for the system to correct prior learning errors and find elements of the excluded groups that may lead to more positive outcomes. So once the model has learned to shut out a particular group, it has little opportunity to unlearn said trend.

## Conclusion

It is often said that data science is both part art and part science. Subjective decision making complements theory-based analysis in practically every data science application. The best we can hope for is that a data scientist develops good judgment and awareness while using the tools at her disposal.

In that sense managing and reducing potential discrimination in a machine learning system follows a similar line of thinking and mechanical processes as one may take when managing and reducing regular statistical biases. Like with statistical bias, the first defense is awareness of how discrimination can enter the system and affect learning objectives.

Our taxonomy of causes highlights exactly where data scientists should look to anticipate discrimination bias. While we do not propose any new methodological techniques, we draw attention to prior research providing such techniques. Our hope is that this survey might serve as a useful first guide in both discrimination measurement and discrimination-aware data mining.

Detection and awareness are the first steps towards enabling practitioners to address the alarming trend of opaque and unaccountable algorithms. As a society, we have much to gain by deploying data for productive means. Data scientists – as stewards of the algorithm – have a unique role to play in this effort. Data scientists share in the ethical and legal obligation to measure and reduce disparate impact induced by algorithms, and to share in the accountability for any potential harm. For the data science program to be successful, we must definitively leave the world a better place, and not merely scale our worst excesses and abuse.